\documentclass[sigconf]{acmart}
\usepackage{tabularx}
\usepackage{subcaption}
\usepackage{multirow}
\usepackage{booktabs}
\usepackage{lipsum}

\copyrightyear{2025}
\acmYear{2025}
\setcopyright{cc}
\setcctype{by}
\acmConference[FDG '25]{International Conference on the Foundations of Digital Games}{April 15--18, 2025}{Graz, Austria}
\acmBooktitle{International Conference on the Foundations of Digital Games (FDG '25), April 15--18, 2025, Graz, Austria}
\acmPrice{}
\acmDOI{10.1145/3723498.3723845}
\acmISBN{/25/04}
\begin{document}

\title{Level Generation with Constrained Expressive Range}

\author{Mahsa Bazzaz}
\affiliation{%
  \institution{Northeastern University}
  \city{Boston, Massachusetts}
  \country{USA}}
\email{bazzaz.ma@northeastern.edu}
\orcid{0009-0004-0022-9611}

\author{Seth Cooper}
\affiliation{%
  \institution{Northeastern University}
  \city{Boston, Massachusetts}
  \country{USA}}
\email{se.cooper@northeastern.edu}
\orcid{0000-0003-4504-0877}

\begin{abstract}
Expressive range analysis is a visualization-based technique used to evaluate the performance of generative models, particularly in game level generation. It typically employs two quantifiable metrics to position generated artifacts on a 2D plot, offering insight into how content is distributed within a defined metric space. In this work, we use the expressive range of a generator as the conceptual space of possible creations. Inspired by the quality diversity paradigm, we explore this space to generate levels. To do so, we use a constraint-based generator that systematically traverses and generates levels in this space. To train the constraint-based generator we use different tile patterns to learn from the initial example levels. We analyze how different patterns influence the exploration of the expressive range. Specifically, we compare the exploration process based on time, the number of successful and failed sample generations, and the overall interestingness of the generated levels. Unlike typical quality diversity approaches that rely on random generation and hope to get good coverage of the expressive range, this approach systematically traverses the grid ensuring more coverage. This helps create unique and interesting game levels while also improving our understanding of the generator’s strengths and limitations.

\end{abstract}

\maketitle

\section{Introduction}
In the field of procedural content generation for game levels \citep{summerville2018procedural}, expressive range has long been the primary tool for qualitative exploration. In procedural content generation via machine learning, expressive range is used to compare the range of a trained content generator to that of the original corpus, with the goal being for the generator to replicate the range properties of the original dataset. Expressive range examines the variety of content that can be created by a generator, according to selected ``metrics'' that can be computed about the content, commonly using these metrics to place content into discrete ``cells'' in a two-dimensional histogram.

Inspired by the quality diversity (QD) paradigm~\citep{10.5555/3692070.3692511}, we highlight the importance of first finding diverse solutions and then improving their quality. This idea comes from the QD principle of rewarding diversity to uncover paths toward high-performing areas in the search space~\citep{Gravina2019ProceduralCG}.

In this work, we use the expressive range of a generator as this conceptual space of potential creations and extend it to explore unique and interesting level designs. To achieve this, we utilize a prioritized random selection to explore underrepresented cells in a two-dimensional expressive range defined by density and difficulty. We employ a constraint-based level generator to create examples from these underrepresented cells. 

By guiding the generator toward unexplored regions of the expressive range, our approach encourages the creation of levels that differ from those typically produced. This not only helps diversify the generated content but also enables a deeper understanding of the generator’s capabilities and limitations. 

The contributions of this paper are:
\begin{itemize}
  \item Utilizing the expressive range space as the conceptual space of possible game creations.
  \item Developing a pipeline for sampling from cells of expressive range space.
  \item Discovering unique and interesting levels by exploring the extended expressive range of the corpus.

\end{itemize}
\section{Related Work}
\begin{figure*}[h]
    \centering
    \includegraphics[width=\textwidth]{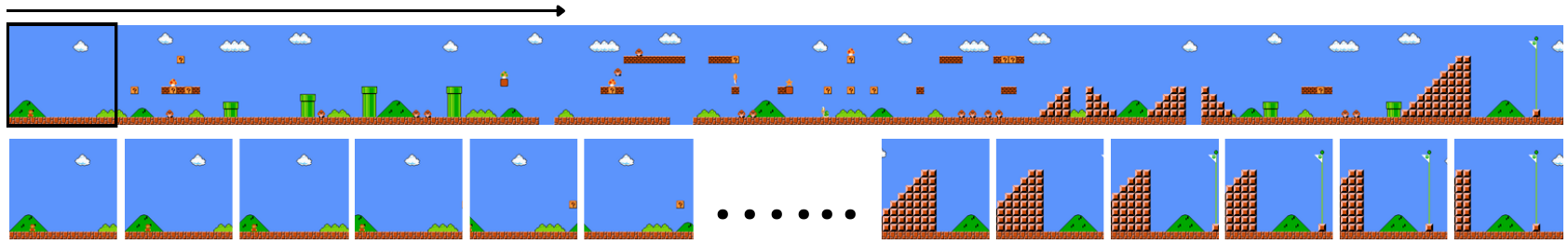}
    \caption{Sliding window horizontally across the level 1-1 of Super Mario Bros. to create the training set.}
    \label{fig:slide}
\end{figure*}
\subsection{Expressive Range Analysis}
The concept of expressive range was first introduced by \citet{10.1145/1814256.1814260} as a means of visualizing generative spaces and assessing the variety and quality of levels produced by generators. They employed metrics such as linearity, leniency, and density to analyze the output tendencies of their system, focusing on characteristics like the types of levels that are likely to be generated and those that are difficult or impossible to produce.

Building on this foundation, \citet{Summerville_2018} introduced corner plots to visualize multiple metric pairs simultaneously, overcoming the previous limitation of only being able to examine two metrics at a time.

Traditional expressive range and corner plots have become fundamental methods for evaluating procedural content generation systems. For instance, \citet{kreminski2022evaluating} utilized expressive range as an evaluation metric to determine how effectively a creative tool supports a wide variety of outputs and how well the interface facilitates both system-initiated and user-driven design choices. Similarly, \citet{alvarez2020interactive} employed expressive range to evaluate the diversity and variety of solutions generated within the feature space of the map-elites algorithm. \citet{digra752} utilized corner plots to illustrate the expressive range of individual metrics along the diagonal and the interplay of metric pairs in the off-diagonal cells. This approach allowed the authors to evaluate whether the generated levels mirrored human-designed levels in specific metrics and preserved a similar overall distribution of characteristics.

The effectiveness of these evaluations, however, can depend on the chosen metrics. There has been some work to enhance this approach (for example \citet{10.1145/3582437.3582453}) by focusing on better capturing meaningful variations in generated content by refining metric selection. By choosing metrics that are relevant and aligned with the intended creative outcomes, it can be possible more accurately and insightfully evaluate PCG systems.

Moreover, other research has tried to define new evaluations that are more dynamic than expressive range. For example, \citet{herve2023exploring} introduced generative shift analysis as a method for evaluating and comparing procedural content generators. This approach emphasizes analyzing shifts in output distributions when the generative system or its parameters are altered, which is in contrast to the expressive range, which measures content variety through static scores.

Expressive range analysis has been used in design in a few studies. For example, during the development of the game That’s Me TV, \citet{10.1145/3402942.3409605} used expressive range visualization to adjust the game's mechanics until they matched his design goals. Similarly, \citet{10.1145/3555858.3555895} used expressive range as a way for designers to directly interact with their probabilistic graph grammar system, allowing them to define their desired generative space by selecting a region on a plot.

\subsection{Quality Diversity}

\begin{figure}[h]
\centering
\begin{tabular}{ccc}
\subcaptionbox{\textit{ring}}[0.10\textwidth]{\includegraphics[width=0.10\textwidth]{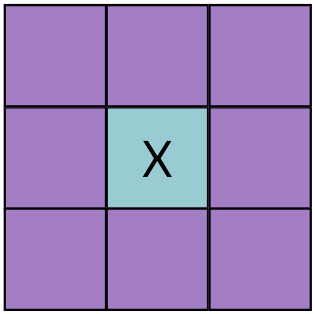}} &
\subcaptionbox{\textit{block2}}[0.10\textwidth]{\includegraphics[width=0.10\textwidth]{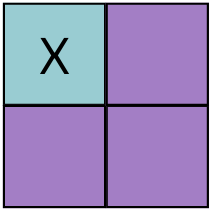}} &
\subcaptionbox{\textit{nbr-plus}}[0.2\textwidth]{\includegraphics[width=0.2\textwidth]{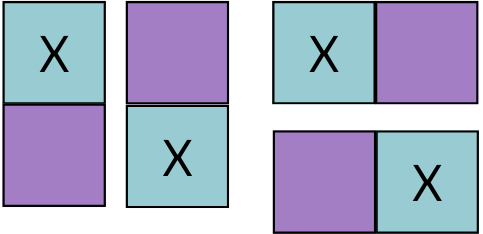}} \\
\end{tabular}
\caption{Pattern templates. The blue tiles are ``input'' tile locations, and the purple are ``output'' tile locations; a certain input tile constrains the corresponding output tiles to be those that were seen in the example level(s). Note that  \texttt{diamond} \texttt{ring} and \texttt{block2} jointly constrain the output tiles all at once, while \texttt{nbr-plus} constrains each output tile independently.}
\label{fig:patterns}
\end{figure}

Quality Diversity (QD) algorithms are optimization methods designed to find many high-quality solutions that are also diverse~\citep{Gravina2019ProceduralCG}. Unlike traditional methods that focus on a single best solution, QD algorithms aim to explore a wide range of possibilities. They reward both quality (how well a solution performs) and diversity (how different the solutions are from each other). By encouraging exploration and ensuring quality through techniques like local competition or constraints, QD algorithms are especially useful for creative tasks~\citep{10.1145/3472538.3472545}.

The expressive range, representing the conceptual space of possible levels that can be generated, creates a partitioned space where diversity can be enforced within a grid of cells. Each cell represents a distinct area in the behavior space, characterized by specific properties, and contains all individuals that exhibit certain behaviors. Then the simplest way to enhance the quality is using local competition between individuals within the same niche.

\section{Methodology}

\subsection{Metric Pair Selection}

We use the two scores of density-difficulty to analyze the 2D expressive range of a corpus of levels. This score combination has been widely adopted in previous works~\citep{sarkar2020sequential,sarkar2020controllable}. We implemented the same definition of these scores from previous work:

\textbf{Density}: the number of tiles in a segment that aren't background tiles.

\textbf{Difficulty}: the number of tiles of a segment that are occupied by any enemy or are hazard tiles (including gaps).

\subsection{Constraint-Based Generator}
\begin{figure*}[!t]
\centering
\begin{tabular}{cccc}
\subcaptionbox{\textit{initial corpus}}[0.24\textwidth]{\includegraphics[width=0.24\textwidth]{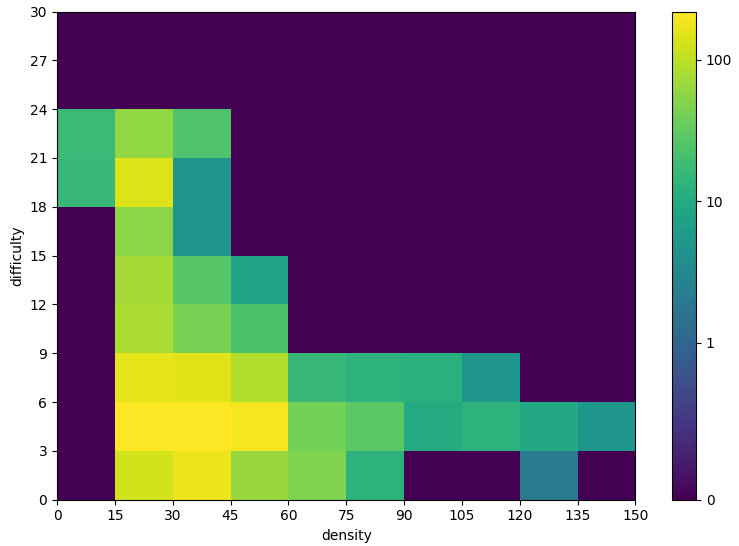}} &
\subcaptionbox{\textit{extended corpus - ring}}[0.24\textwidth]{\includegraphics[width=0.24\textwidth]{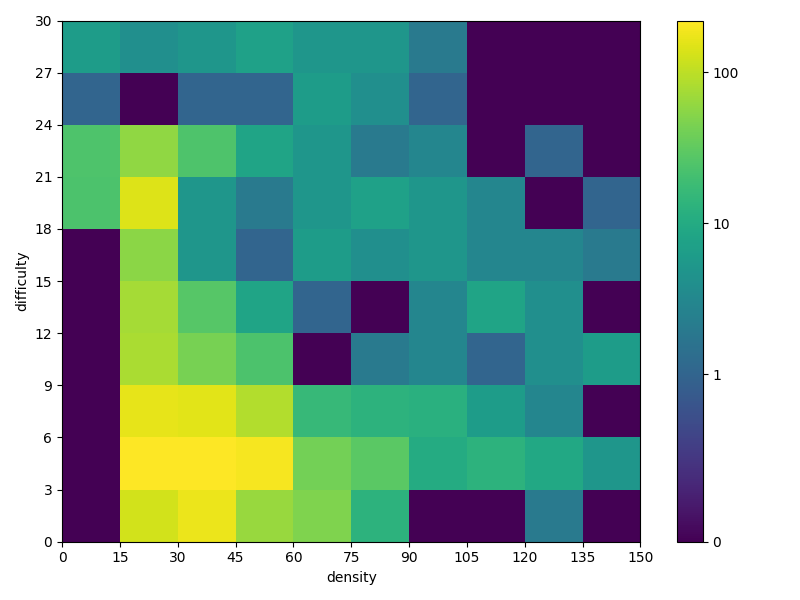}} &
\subcaptionbox{\textit{extended corpus - block2}}[0.24\textwidth]{\includegraphics[width=0.24\textwidth]{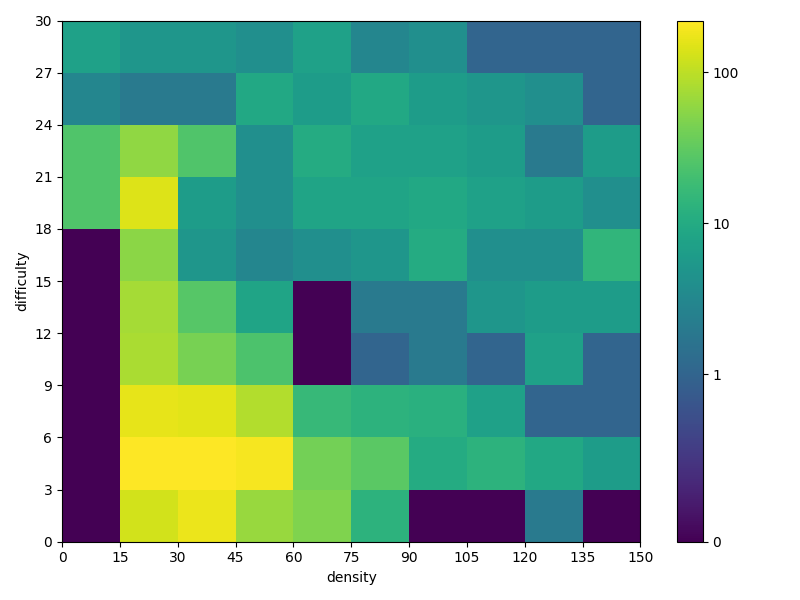}} &
\subcaptionbox{\textit{extended corpus - nbr-plus}}[0.24\textwidth]{\includegraphics[width=0.24\textwidth]{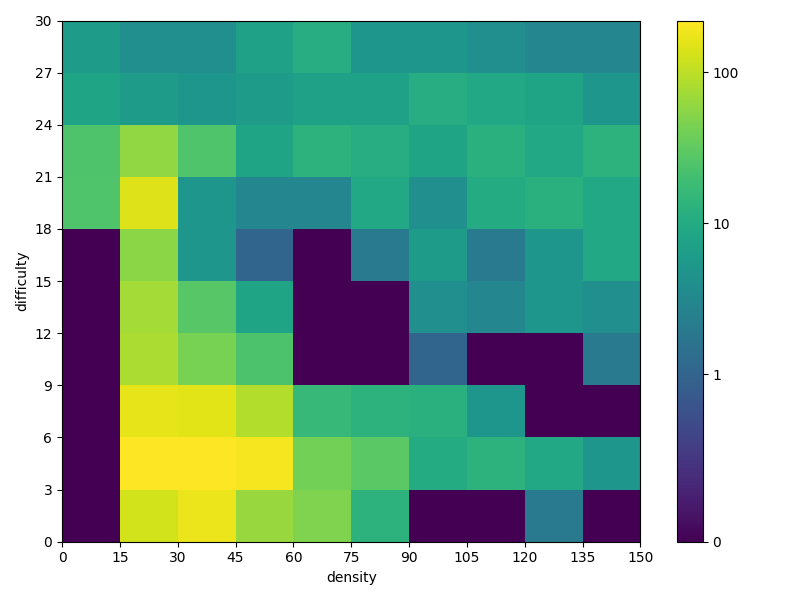}} \\
\end{tabular}
\caption{Density-difficulty expressive range of expanded corpus with each pattern template. The color bar is based on \textbf{SymLogNorm} allowing linear behavior around zero and logarithmic beyond.}
\label{fig:results}
\end{figure*}

\begin{figure*}[!h]
\centering
\begin{tabular}{cccc}
\subcaptionbox{\textit{initial corpus}}[0.24\textwidth]{\includegraphics[width=0.24\textwidth]{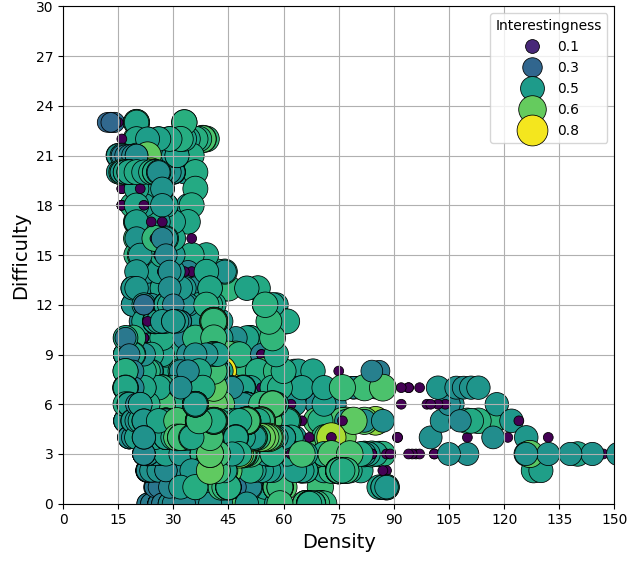}} &
\subcaptionbox{\textit{extended corpus - ring}}[0.22\textwidth]{\includegraphics[width=0.22\textwidth]{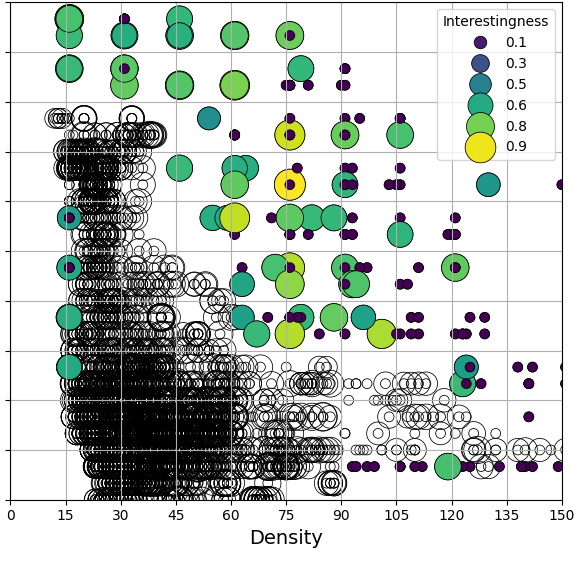}} &
\subcaptionbox{\textit{extended corpus - block2}}[0.22\textwidth]{\includegraphics[width=0.22\textwidth]{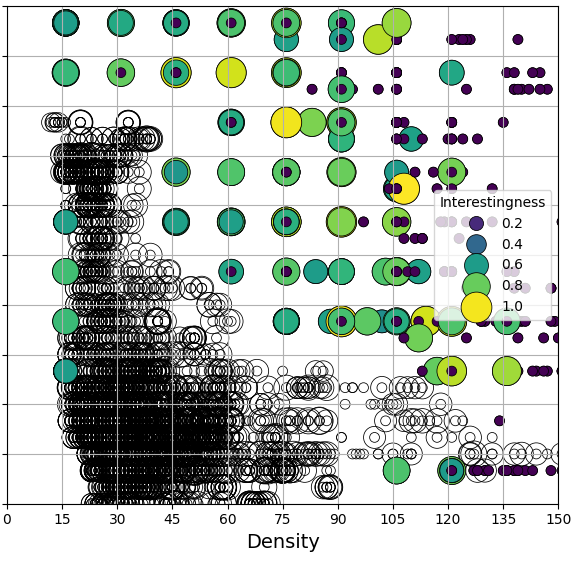}} &
\subcaptionbox{\textit{extended corpus - nbr-plus}}[0.22\textwidth]{\includegraphics[width=0.22\textwidth]{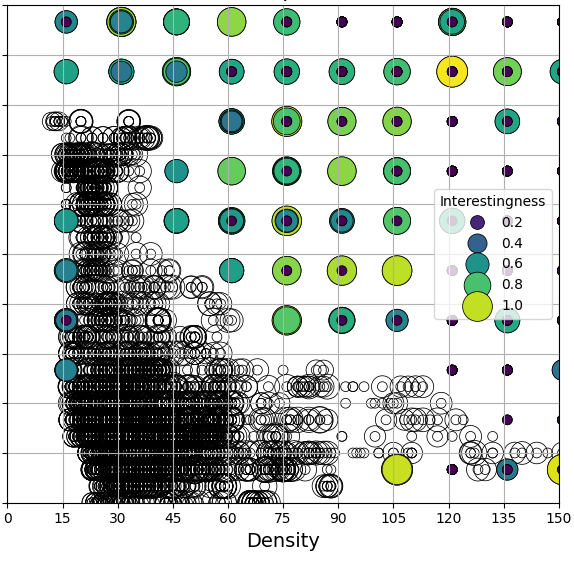}} \\
\end{tabular}
\caption{Scatter plot displaying the (normalized) interestingness of generated levels across the expressive range, with the size and color of each point reflecting its level of interestingness. Uncolored points represent levels from the initial corpus in extended corpuses.}
\label{fig:interestingness}
\end{figure*}

\begin{table*}[!h]
    \centering
    \resizebox{\textwidth}{!}{%
    \begin{tabular}{|c|c|c|c|c|c|c|c|}
        \hline
         \textbf{Pattern Template} & \textbf{Total Attempts} & \textbf{Successful Attempts} & \textbf{Failed Attempts} & \textbf{Timed Out Attempts} & \textbf{Average Solve Time (s)} & \textbf{Average Fail Time (s)} & \textbf{Average Time (s)} \\ 
        \hline
         \texttt{ring} & 200 & 177 & 2 & 21 & 150.09 & 73.73 & 374.61 \\
         \texttt{block2} & 294 & 282 & 2 & 10 & 8.36 & 19.97 & 309.44 \\
         \texttt{nbr-plus} & 315 & 313 & 2 & 0 & 0.98 & 30.78 & 10.57 \\
        \hline
    \end{tabular}%
    }
    \caption{Comparison of performance of different pattern templates in terms of successful attempts (resulted in level generation) within a fixed time limit.}
    \label{tab:times}
\end{table*}

In this work, we use the Sturgeon constraint-based level generator \citep{Cooper_2022} as a dependable tool for procedural generation. Sturgeon leverages a straightforward, mid-level API to define constraints on Boolean variables, which it translates into low-level constraint satisfaction problems. As a result, Sturgeon ensures that the generated levels meet the specified constraints precisely, without any errors or noise.

Sturgeon enforces local tile patterns as constraints, making sure the generated levels replicate the local structures of example levels. This is achieved by using pattern templates that create levels by using rules that guide how tiles should be arranged, rather than placing them randomly. Each pattern template consists of a set of input and output patterns, where specific input patterns limit the types of output patterns that can appear. In this work, we experimented with different pattern templates (\texttt{ring}, \texttt{block2}, and  \texttt{nbr-plus}; described below) to analyze their performance in exploring the generator's expressive range.

For this work, we used Super Mario Bros.~\citep{GAME_mariobros} levels from The VGLC~\citep{VGLC} to extract these patterns. Additionally, Sturgeon can impose constraints on overall tile counts, allowing the generated levels to maintain a similar overall appearance to the example. We also utilized these constraints on overall tile counts to enforce specific metric values to sample from the 2D expressive range of possible game creations.

\section{Experiments}
\subsection{Domain}
To create the initial corpus of Super Mario levels, we applied a sliding window of size \mbox{$20\times14$} horizontally across the level, producing $2,302$ smaller segments with  $11$ tile types. Figure \ref{fig:slide} shows a visualization of this process. 

\subsection{Exploring the Expressive Range Space}

To apply Sturgeon's capability of imposing constraints on overall tile counts, we need to frame the density-difficulty constraint as a tile count problem for Sturgeon. This is straightforward for both density and difficulty metrics, as they are computed based on tile counts. We aim to satisfy the constraint that the total number of non-background tiles equals the desired density, while the sum of enemy and hazard tiles corresponds to the desired difficulty.

\begin{figure*}[h]
\centering
\resizebox{0.95\textwidth}{!}{%
\begin{tabular}{cccc}
\subcaptionbox{\textit{initial}}[0.24\textwidth]{\includegraphics[width=0.24\textwidth]{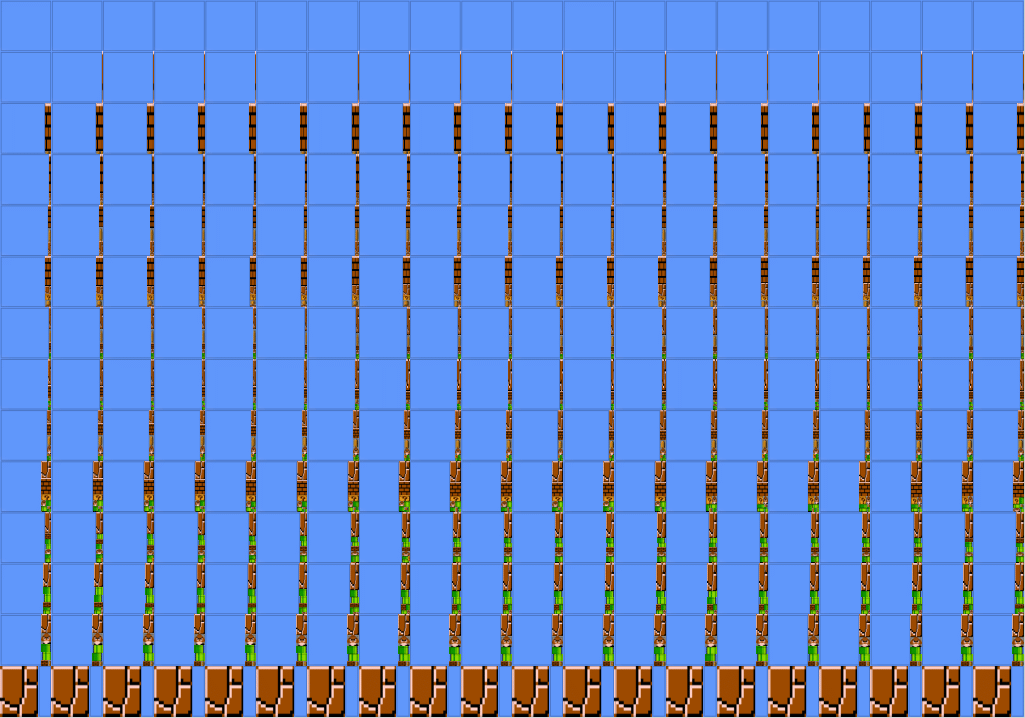}} &
\subcaptionbox{\textit{ring}}[0.24\textwidth]{\includegraphics[width=0.24\textwidth]{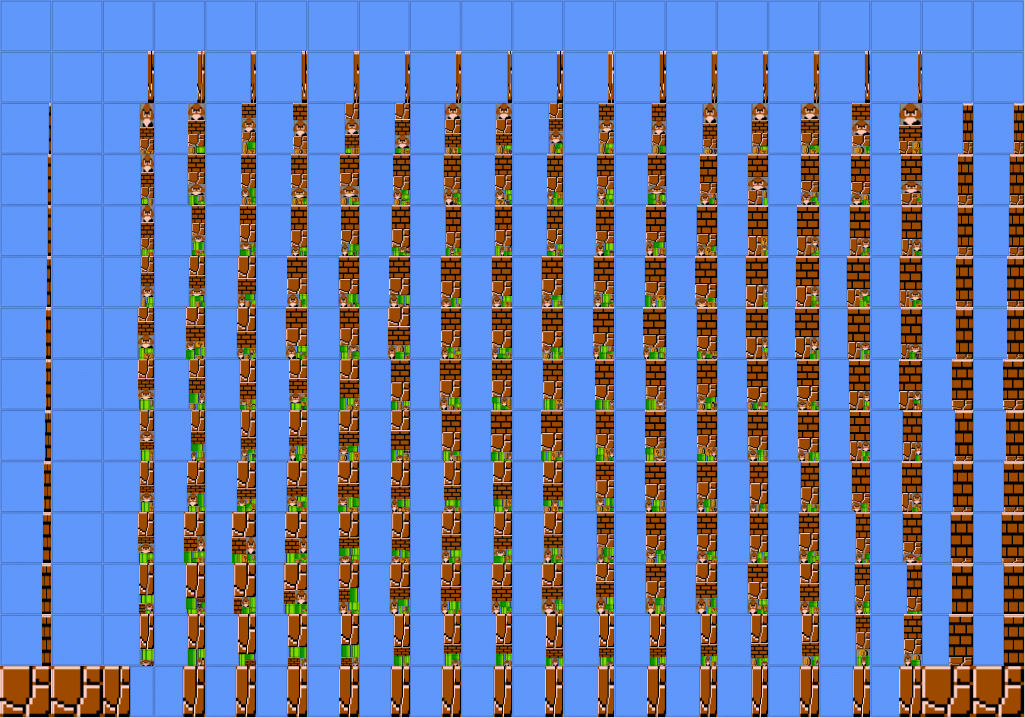}} &
\subcaptionbox{\textit{block2}}[0.24\textwidth]{\includegraphics[width=0.24\textwidth]{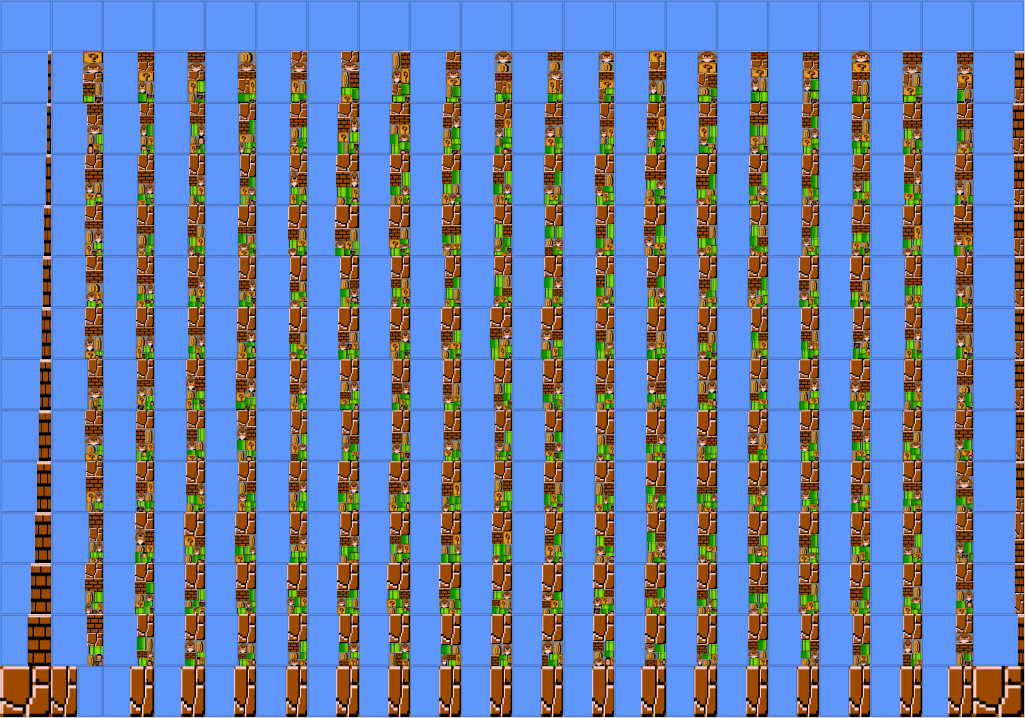}} &
\subcaptionbox{\textit{nbr-plus}}[0.24\textwidth]{\includegraphics[width=0.24\textwidth]{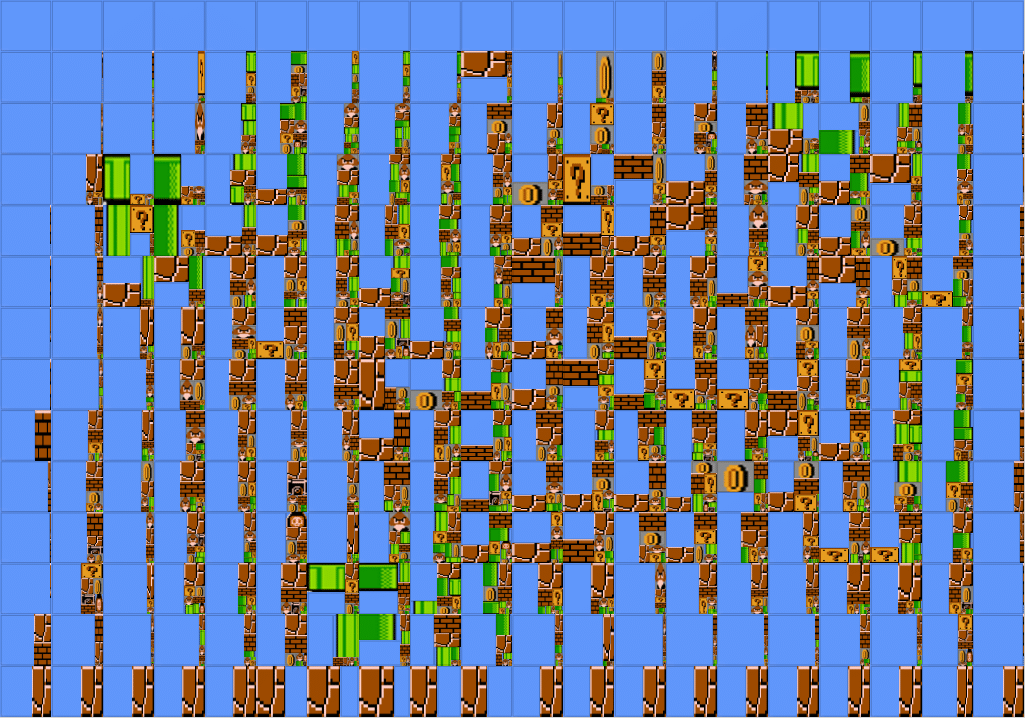}} \\
\end{tabular}
}
\caption{Visual exploration of initial levels of the corpus and the generated levels with different pattern templates.}
\label{fig:visual}
\end{figure*}

In this work, we experimented with the \texttt{ring}, \texttt{block2}, and \texttt{nbr-plus} pattern templates. Figure \ref{fig:patterns} illustrates the \texttt{ring} pattern template is the most constrained template between these three followed by \texttt{block2} and then \texttt{nbr-plus}. More constrained pattern templates (e.g. \texttt{ring}) impose greater structure on level generation, making it harder to fill gaps and create complete levels. Therefore, a trade-off exists between maintaining structure and finding feasible solutions when selecting different pattern templates.

To compare the different pattern templates based on their speed in exploring the expressive range space, we allocated the same amount of time (12 hours) for all three pattern templates. 
The exploration process uses a prioritized random selection that focuses on underrepresented cells within the space --- cells containing fewer than 10 levels. At each step, cells with the fewest levels are selected from uniformly at random. Underrepresented cells can be selected multiple times until they reach the 10-level threshold. If an attempt to generate a level in a cell fails, or if the cell has already reached the limit, it is added to a blocklist. This prevents the process from getting stuck in cells that are either impossible to generate or have already met their level count.

A pattern template that successfully generates a level meeting the density-difficulty constraint, or identifies the absence of solutions for that setting, can explore the space more efficiently. In addition to the overall allocated time, we have set a 15-minute timeout for individual level generation attempts to prevent solvers from getting stuck on more challenging density-difficulty settings.

This approach is somewhat similar to reversing the typical flow of quality diversity searches. Instead of generating levels and hoping they cover a wide expressive range, we systematically explore the space to ensure full coverage. To investigate this further, we also run the constraint-based generator without any specific constraints, allowing it to create levels randomly, without considering density or difficulty. We then compare how well these purely random levels cover the density-difficulty space.

To assess the quality of the generated levels, we introduce a quantitative measure of ``interestingness''. Using Sturgeon, we determine the player’s path in each generated segment and examine two key factors: path length (which reflects gameplay duration) and the number of jumps (which reflects the player's engagement). By adding these values together, we estimate how interesting the levels in each explored cell are. This quantitative measure is inspired by gameplay features detected in previous works, that significantly correlate to being ``fun''\citep{5286482}. 

The definition of ``interestingness'' is not a definitive measure of the quality of the generated levels, but it provides insight into which levels are engaging and playable, rather than bland or unplayable.

\section{Results}
\subsection{Exploring the Expressive Range Space}
As anticipated, the various template patterns demonstrated differing performances in exploring the expressive range of games. Figure \ref{fig:results} illustrates how effectively each pattern explored the space. The \texttt{ring} pattern, due to its more restrictive constraints, was less successful in covering all cells. However, as the patterns relaxed these constraints, their ability to find level solutions for each cell improved.

As shown in Table \ref{tab:times}, the \texttt{nbr-plus} pattern template achieved the lowest solve time, while the \texttt{ring} pattern template had the highest. This outcome supports the expectation that less-constrained pattern templates are faster at finding solutions, highlighting the trade-off between maintaining an ideal structure and optimizing speed. This trade-off is further evident in the samples generated during the exploration of the expressive range, as shown in Figure \ref{fig:sturgeon_samples}. 

\begin{figure*}[h]
\centering
\begin{tabular}{cccc}
\subcaptionbox{\textit{initial}}[0.24\textwidth]{\includegraphics[width=0.24\textwidth]{figs/er/mario_expressive_range.png}} &
\subcaptionbox{\textit{ring}}[0.24\textwidth]{\includegraphics[width=0.24\textwidth]{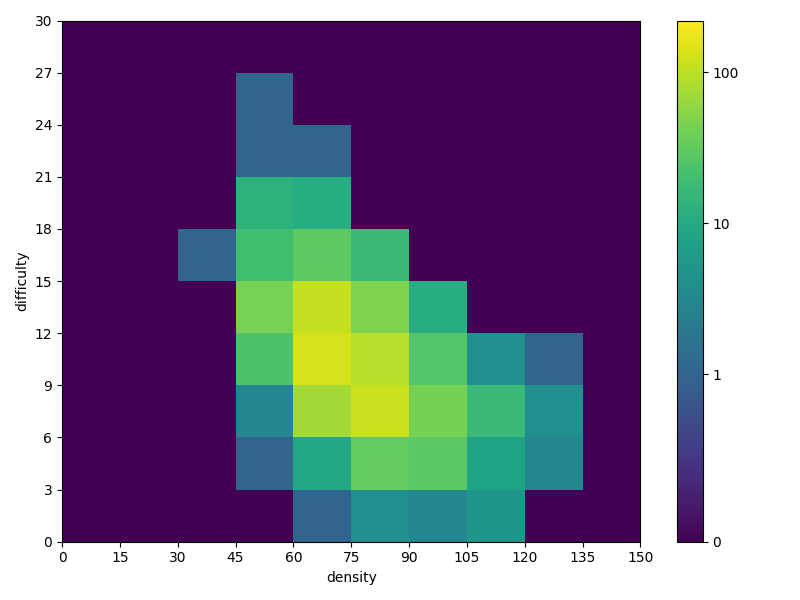}} &
\subcaptionbox{\textit{block2}}[0.24\textwidth]{\includegraphics[width=0.24\textwidth]{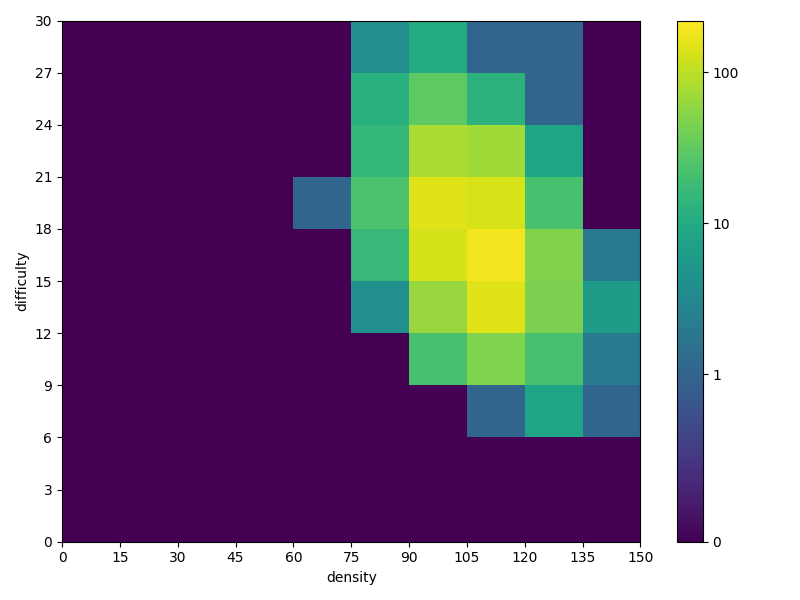}} &
\subcaptionbox{\textit{nbr-plus}}[0.24\textwidth]{\includegraphics[width=0.24\textwidth]{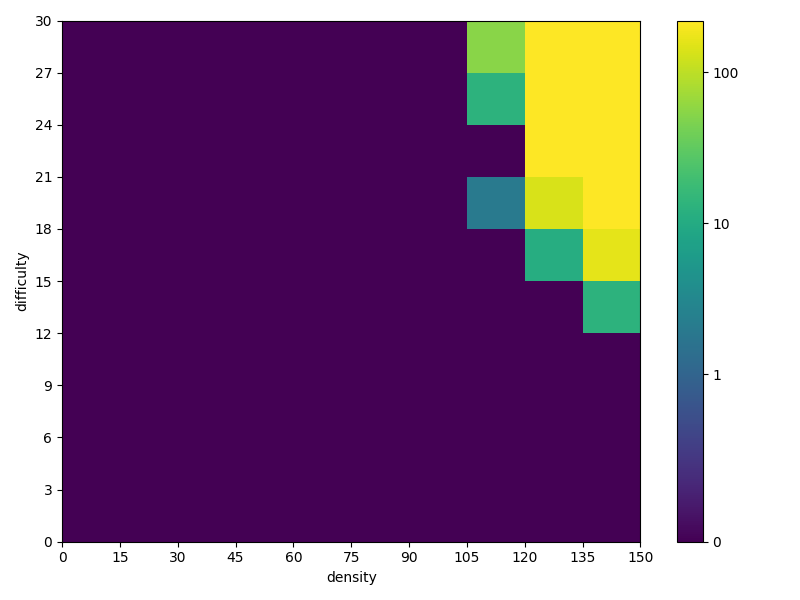}} \\
\end{tabular}
\caption{Density-difficulty expressive range of RANDOM generated corpus with each pattern template. The color bar is based on \textbf{SymLogNorm} allowing linear behavior around zero and logarithmic beyond.}
\label{fig:results_random}
\end{figure*}

Comparing the expanded expressive range of the systematic traversal in Figure \ref{fig:results} with the fully random level generation in Figure \ref{fig:results_random} highlights a significant difference in coverage between these two approaches. Interestingly, random generation for each pattern template tends to cover entirely different areas of the expressive range. This is likely because different pattern templates are better suited to capturing specific structures in levels, making them more effective for generating different densities and difficulties.

\begin{figure*}[h]
\centering
\begin{tabular}{cccc}
\subcaptionbox{\textit{initial}}[0.24\textwidth]{\includegraphics[width=0.24\textwidth]{figs/new/initial-random.png}} &
\subcaptionbox{\textit{ring}}[0.22\textwidth]{\includegraphics[width=0.22\textwidth]{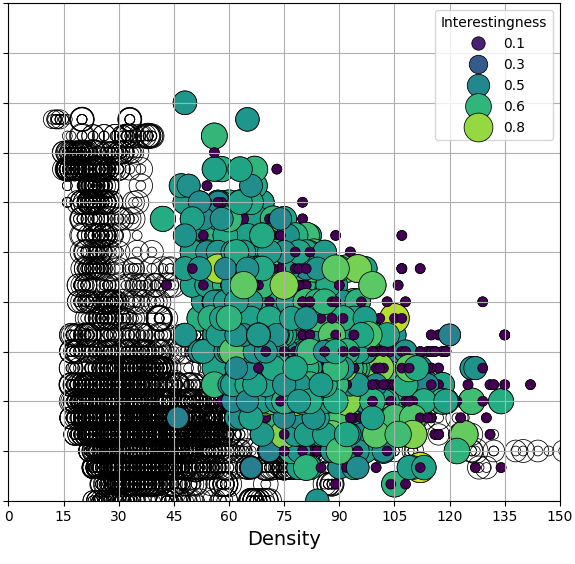}} &
\subcaptionbox{\textit{block2}}[0.22\textwidth]{\includegraphics[width=0.22\textwidth]{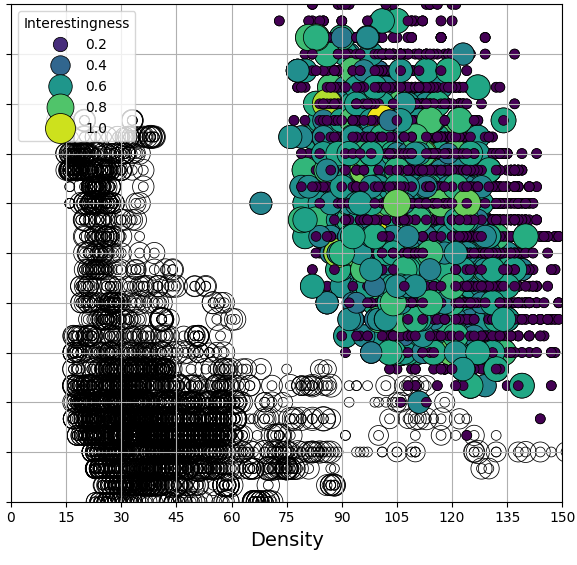}} &
\subcaptionbox{\textit{nbr-plus}}[0.22\textwidth]{\includegraphics[width=0.22\textwidth]{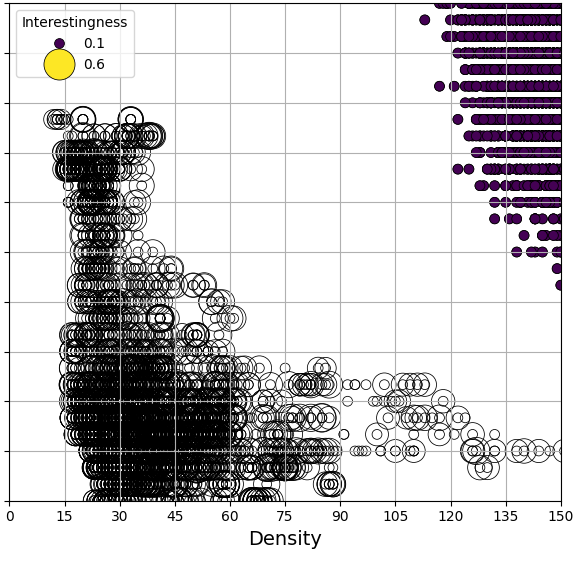}} \\
\end{tabular}
\caption{Scatter plot displaying the (normalized) interestingness of complete randomly generated levels across the expressive range, with the size and color of each point reflecting its level of interestingness. uncolored points represent levels from the initial corpus in extended corpuses.}
\label{fig:interestingness_random}
\end{figure*}

Additionally, we compare the (normalized) interestingness of levels generated through systematic traversal (Figure \ref{fig:interestingness}) with those created via random generation (Figure \ref{fig:interestingness_random}). The interestingness values have been normalized between the minimum and maximum values across all values to make sure the figures are easily comparable.  We observe some differences. For instance, it is clear that the \texttt{nbr-plus} pattern template struggles to produce levels that are both playable and interesting without guidance on the appropriate proportions of density and difficulty --- or, more broadly, without additional constraints on the level.

\subsection{Visual Exploration}

We used the visual explorer tool designed by \citet{cooper2023visual} to provide a quick visual summary of both the initial and extended corpora. Inspired by Oskar Stålberg \citep{Stalberg2018}, this tool represents all available tiles within each cell, scaled according to their frequency. Comparing the visual exploration of the initial levels of the corpus and the generated levels with different pattern templates in Figure \ref{fig:visual} shows some immediate differences between the levels that come from underrepresented cells of expressive range and the rest. For example, the levels generated for underrepresented cells show noticeable variations in pipe positions and heights. This trend is more noticeable in patterns that are less restrictive (like \texttt{nbr-plus}) as they allow more uncommon structures in levels.

\section{Discussion}

This work is an initial attempt to create a systematic method that allows Sturgeon to explore its expressive range in a controlled manner, using an approach inspired by quality diversity, but with full control over the exploration. In the future, we plan to build on this by adding more complex metrics, like linearity, which go beyond counting tiles and are harder to define as a satisfaction problem for the generator.

We believe that this controlled exploration of the expressive range can offer value in the context of generative model training. By curating specific datasets through this approach, we can ensure that the data used for training is more diverse and tailored to particular characteristics or qualities.

\section{Conclusion}
In this work, we use the expressive range of a level generator as a conceptual space to explore potential level designs. To investigate underrepresented areas in terms of density and difficulty, we employed a prioritized random selection strategy with a constraint-based generator. By applying different pattern templates to learn from initial levels, we evaluated the efficiency of level generation based on time, success and failure rates, and the interestingness of the generated levels. This approach allows us to interact with the generator, covering the expressive range systematically (rather than randomly) to produce diverse and interesting game levels.

\begin{acks}
Support was provided by Research Computing at Northeastern University (\url{https://rc.northeastern.edu/}) through the use of the Discovery Cluster.
\end{acks}

\bibliographystyle{ACM-Reference-Format}
\bibliography{refs}

\appendix
\section{Appendix}

\begin{figure}[b!]
\centering
\resizebox{0.52\textwidth}{!}{%
\begin{tabular}{ccc}
\subcaptionbox*{\textit{ring}}[0.3\textwidth]{\includegraphics[width=0.3\textwidth]{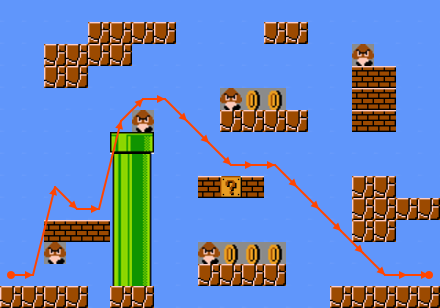}} &
\subcaptionbox*{\textit{block2}}[0.3\textwidth]{\includegraphics[width=0.3\textwidth]{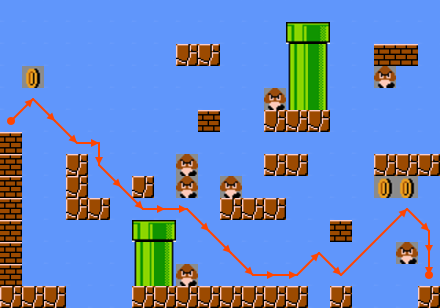}} &
\subcaptionbox*{\textit{nbr-plus}}[0.3\textwidth]{\includegraphics[width=0.3\textwidth]{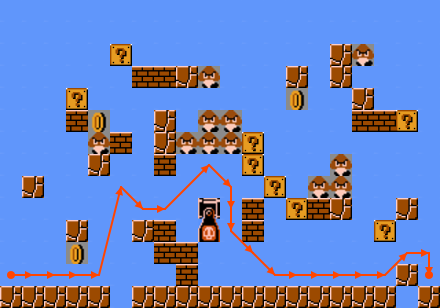}} \\
\multicolumn{3}{c}{\subcaptionbox{density = (60-75), difficulty = (12-15)}[0.6\textwidth]{}} \\
\subcaptionbox*{\textit{ring}}[0.3\textwidth]{\includegraphics[width=0.3\textwidth]{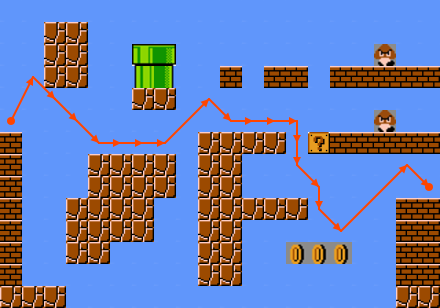}} &
\subcaptionbox*{\textit{block2}}[0.3\textwidth]{\includegraphics[width=0.3\textwidth]{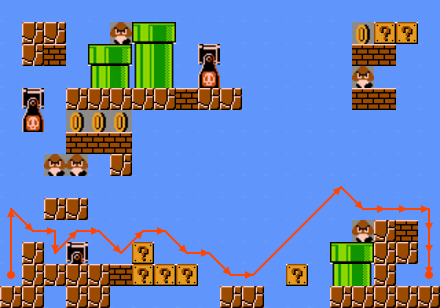}} &
\subcaptionbox*{\textit{nbr-plus}}[0.3\textwidth]{\includegraphics[width=0.3\textwidth]{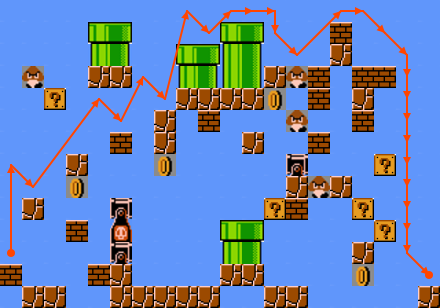}} \\
\multicolumn{3}{c}{\subcaptionbox{density = (75-90), difficulty = (15-18)}[0.6\textwidth]{}} \\
\subcaptionbox*{\textit{ring}}[0.3\textwidth]{\includegraphics[width=0.3\textwidth]{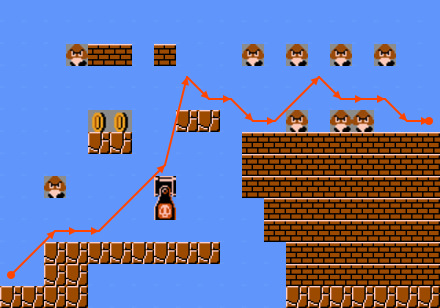}} &
\subcaptionbox*{\textit{block2}}[0.3\textwidth]{\includegraphics[width=0.3\textwidth]{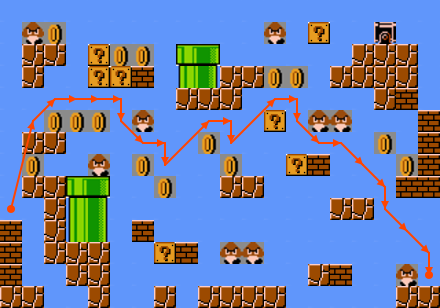}} &
\subcaptionbox*{\textit{nbr-plus}}[0.3\textwidth]{\includegraphics[width=0.3\textwidth]{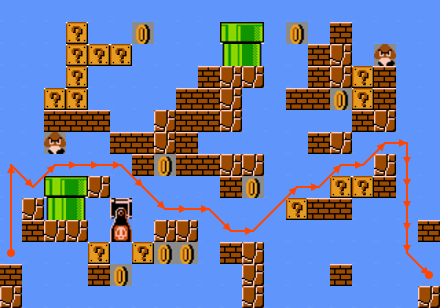}}\\
\multicolumn{3}{c}{\subcaptionbox{density = (90-105), difficulty = (18-21)}[0.6\textwidth]{}} \\
\subcaptionbox*{\textit{ring}}[0.3\textwidth]{\includegraphics[width=0.3\textwidth]{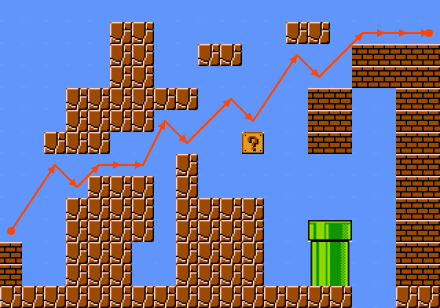}} &
\subcaptionbox*{\textit{block2}}[0.3\textwidth]{\includegraphics[width=0.3\textwidth]{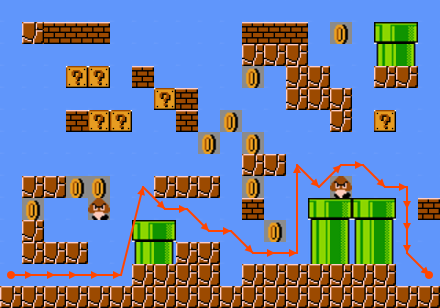}} &
\subcaptionbox*{\textit{nbr-plus}}[0.3\textwidth]{\includegraphics[width=0.3\textwidth]{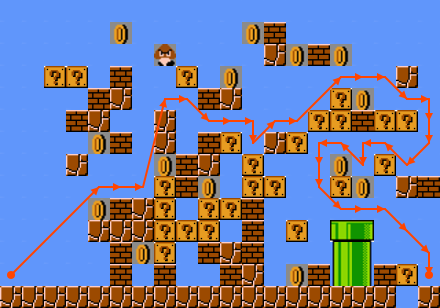}}\\
\multicolumn{3}{c}{\subcaptionbox{density = (105-120), difficulty = (0-3)}[0.6\textwidth]{}} \\
\end{tabular}
}
\caption{Samples of generated levels with constraint-based level generator (Sturgeon) while exploring the same underrepresented cells in expressive range space with different pattern templates.}
\label{fig:sturgeon_samples}
\end{figure}

\end{document}